\documentclass{article}
\pdfoutput=1
\usepackage[letterpaper,margin=1in,rmargin=1in,lmargin=1in]{geometry}
\usepackage[utf8]{inputenc}

\usepackage{kpfonts}
\usepackage[T1]{fontenc}
\usepackage{authblk}

\usepackage{hyperref}
\usepackage{url}

\usepackage{amsmath}
\usepackage{amssymb}
\usepackage{xspace}
\usepackage{graphicx}
\usepackage{color}
\usepackage{bm}
\usepackage{bbm}
\usepackage[caption=false]{subfig}
\usepackage{wrapfig}
\usepackage{booktabs}

\usepackage{algorithm}
\usepackage{algorithmic}

\usepackage{microtype}
\usepackage{booktabs}

\usepackage{comment}
\usepackage{tabularx,ragged2e,booktabs}
\usepackage{adjustbox}
 \usepackage{relsize}
\usepackage{rotating}
\usepackage{multirow}
\usepackage{diagbox}
\usepackage{stmaryrd}
\usepackage{changepage}
\usepackage{placeins}
\usepackage{breqn}
\usepackage{makecell}
\usepackage[inline]{enumitem}
\usepackage{cleveref}
\usepackage{mathtools}
\usepackage[switch]{lineno}
\usepackage[detect-none]{siunitx}
\DeclareUnicodeCharacter{2212}{-}

\DeclareMathOperator*{\E}{\mathbb{E}}
\DeclareMathOperator{\Lagr}{\mathcal{L}}

\newcommand\norm[1]{\left\lVert#1\right\rVert}
\newcolumntype{C}[1]{>{\Centering}m{#1}}
\DeclareMathAlphabet{\mathpzc}{T1}{pzc}{m}{it}

\DeclarePairedDelimiter\ceil{\lceil}{\rceil}
\DeclarePairedDelimiter\floor{\lfloor}{\rfloor}

\newcommand{\at}[0]{AT\xspace}
\newcommand{\fedavg}[0]{\texttt{FedAvg}\xspace}
\newcommand{\fedcurv}[0]{\texttt{FedCurv}\xspace}
\newcommand{\fedavgat}[0]{\texttt{FedAvgAT}\xspace}
\newcommand{\fedcurvat}[0]{\texttt{FedCurvAT}\xspace}

\newcommand{\feddynat}[0]{\texttt{FedDynAT}\xspace}
\newcommand{\fedavgdynat}[0]{\texttt{FedAvgDynAT}\xspace}

\title{Adversarial training in communication constrained federated learning}
\author[1]{Devansh Shah}
\author[2]{Parijat Dube}
\author[2]{Supriyo Chakraborty}
\author[2]{Ashish Verma}

\affil[1]{Department of Computer Science, University of Illinois, Urbana Champaign}
\affil[2]{I.B.M T.J. Watson Research Center}

\date{}

\begin{document}
\setlength{\abovedisplayskip}{0pt}
\setlength{\belowdisplayskip}{0pt}
\setlength{\abovedisplayshortskip}{0pt}
\setlength{\belowdisplayshortskip}{0pt}
\maketitle


\begin{abstract}
    Federated learning enables model training over a distributed corpus of agent data. However, the trained model is vulnerable to adversarial examples -- designed to elicit misclassification. 
    We study the feasibility of using adversarial training (AT) in the federated learning setting. Furthermore, we do so assuming a fixed communication budget and non-iid data distribution between participating agents. 
    We observe a significant drop in both natural and adversarial accuracies when AT is used in the federated setting as opposed to centralized training. We attribute this to the number of epochs of AT performed locally at the agents, which in turn effects (i) \emph{drift} between local models; and (ii) convergence time (measured in number of communication rounds). Towards this end, we propose \feddynat -- a novel algorithm for performing AT in federated setting.
    Through extensive experimentation we show that \feddynat significantly improves both natural and adversarial accuracy, as well as model convergence time by reducing the model drift.
\end{abstract}
\section{Introduction}
\label{sec:intro}

Federated learning is a paradigm for multi-round model training over a distributed dataset~\cite{mcmahan2016communicationefficient, kairouz:open}. At the beginning of every round, the participating clients obtain the most recent model from a central server. Clients use the model to perform multiple epochs of local training and share only the parameter updates with the server. The server, in turn, employs a \textit{fusion} algorithm to aggregate the updates from the various clients and generate a new model for the next round. While federation allows the clients to retain control and privacy of their raw data~\cite{bonawitz:secure, truex:privacy}, the final model continues to remain vulnerable to carefully crafted adversarial examples~\cite{szegedy:intriguing, carlini:c_and_w, goodfellow2015explaining} designed to elicit misclassification at inference time. Complementary to our research, there exists a significant body of work on defenses (both empirical~\cite{madry2018towards} and certifiable~\cite{raghunathan:certify, cohen:smoothing}) against adversarial attacks for centralized model training. Among them, adversarial training~\cite{madry2018towards} (\at), formulated as an optimization of the \emph{saddle point problem}, has emerged as the method of choice for training robust models. In this paper, \emph{we study the feasibility of adopting AT to the federated learning setting for training robust models, under realistic constraints on both the client data distribution (non-iid) and the maximum number of possible communication rounds.}

\begin{figure}[!ht]
\centering
\includegraphics[scale=0.4]{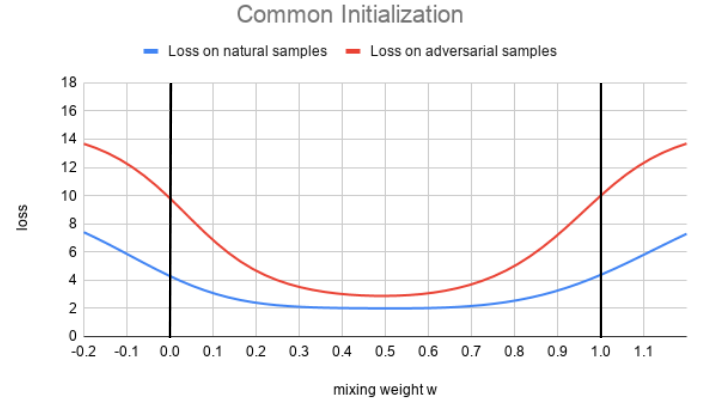} 
\caption{\centering \small{Loss on CIFAR-10 training set (natural and adversarial samples) using aggregate models. The aggregate model parameters $\theta = w \theta_1 + (1 − w) \theta_2$, where $\theta_1$ and $\theta_2$ are client model parameters, with common initialization but updated independently using local \at at the clients. Mixing weights $w \in [−0.2, 1.2]$.}}
\label{fig:feasibility}
\end{figure}

To optimize the minimax objective, \at typically uses multiple iterations of a two-step sequential process: (i) Use Projected Gradient Descent (PGD) to maximize the loss and generate adversarial counterparts for every sample in the training data; (ii) Use Stochastic Gradient Descent (SGD) to optimize the model parameters and minimize the expected loss over the adversarial examples. From an optimization perspective, an immediate consequence of adopting AT to the federated setting is a distributed solution to the minimax objective. In other words, the above two steps, which were earlier performed over the entire data, are now distributed across multiple clients and performed on their local data. Our goal is to design a fusion algorithm, which over multiple rounds ensures that the aggregate model has \emph{comparable} performance in terms of both natural accuracy (measured on benign test samples) and adversarial accuracy (measured on adversarial test samples) w.r.t the model trained in the centralized setting. 
As a first step towards feasibility analysis, we reproduce the observation in  ~\cite{mcmahan2016communicationefficient} for adversarial training. We perform weighted averaging of adversarially trained models from two clients. The models are trained on an iid distributed subset of CIFAR10 data, starting from a common initialization. 
The result shown in Fig.~\ref{fig:feasibility} provides initial evidence that the aggregate model leads to a significant reduction in the loss on the entire training set for both natural and adversarial samples compared to either client models (extreme values of $w$). This implies that fusion can also improve adversarial accuracy (\fedavg had established that fusion helps for natural training).

\noindent{\bf Challenges:} However, there are several challenges that need to be addressed. Using centralized clean training as baseline, we observed a higher drop in both natural and adversarial accuracies when \at is used in the federated setting as opposed to centralized training. We attribute this to number of epochs of local AT, denoted by $E$, performed at each client, which in turn causes the local models at the clients to \textit{drift} further from each other. This is over and above the drift that is already introduced by the non-iid data distribution at the clients. We use SVCCA~\cite{raghu:svcca} scores to quantify model drift across clients.  In addition, the convergence time (in terms of the number of communication rounds) is also increased. This is of particular concern as it increases the communication cost significantly.

\noindent{\bf Contributions:}
Towards this end, we propose \feddynat, which builds on the techniques proposed for preventing catastrophic forgetting in federated learning on non-iid data~\cite{shoham2019overcoming, kirkpatrick2016overcoming} by augmenting them with a dynamic local AT schedule ($E$-schedule). We make the following contributions..

\begin{itemize}[noitemsep,nolistsep]
    \item We study the performance of adversarial training in a federated setting with non-iid data distribution and discuss the challenges involved as compared to normal training.
    \item We present a novel algorithm for improving the performance of adversarial training in federated setting with non-iid data distribution, called \feddynat, and evaluate its performance over multiple datasets and models. We compare the performance with other state-of-the art methods used in normal federated learning.
    \item We analyze the performance of different algorithms in the federated setting using model drift across various clients and provide insights on the relative performance.
\end{itemize}

\section{Adversarial Training in Federated Setting}

Following~\cite{madry2018towards}, we define centralized adversarially trained model parameters,  $\hat{\theta}$, as a solution to the following min-max problem:
 \begin{equation}
 \label{eqn:fedat0}
 \min_{\theta} \rho(\theta); \ \ \ 
 \rho(\theta) =  \E_{(x,y) \thicksim {\cal F}} [\max_{\norm{\Tilde{x}-x}_{\infty} \leq \epsilon} \Lagr(f_{\theta}(\Tilde{x}), y)]
 \end{equation}
 where $f_{\theta}$ is the model parameterized by $\theta$ and $\Tilde{x}$ is an $l_{\infty}$ adversarial sample, within $\epsilon$-ball of the benign sample $x$, with class label $y$.
 
 \noindent{\bf Problem Formulation:} We consider adversarial training between $K$ clients, each having their share of the training data $D_k\thicksim{\cal F}_k$ and $D (=\bigcup_{k=1}^K D_k) \thicksim {\cal F}$.
 Our objective is to train a joint machine learning model {\it without any exchange} of training data between clients.
 Motivated by our initial observations in (Figure~\ref{fig:feasibility}) we investigate adversarial training in federated environment as a combination of local adversarial training at individual clients followed by periodic synchronization and aggregation of model parameters at a trusted server. The aggregation, achieved by {\it fusing} the model weights of individual clients, results in a {\it global model}. The expectation is that after a sufficiently large number of aggregation rounds $t_0$,  this global model  will be an approximate solution to \eqref{eqn:fedat0}, i.e., $\rho(\theta^{t_0}) \approx \rho(\hat{\theta})$, providing robustness to adversarial attacks at all clients. 
 
Let $\theta_k^{t} \in \mathbb{R}^{n}$ be the model parameters at client $k = 1, \ldots, K$ at the beginning of round $t$. The aggregation of model weights is achieved using a fusion operator $F:\mathbb{R}^{K \times n} \rightarrow \mathbb{R}^{n}$. We denote the fusion output at the end of round $t$ as $\theta^{t+1}$. A training round $t$ involves $E_t$ epochs of adversarial training using local data at each agent. Thus, $\theta_k^t$ is a solution to the following min-max problem obtained over $E_t$ epochs with initial value $\theta_k^{t}=\theta^{t}$:
\begin{eqnarray}
\min_{\theta} \rho_k(\theta); \ \rho_k(\theta) & = &  \E_{(x,y) \in D_{k}} [\max_{\norm{\Tilde{x}-x}_{\infty} \leq \epsilon} \Lagr(f_{\theta}(\Tilde{x}), y)] \nonumber \\ 
\mbox{ and } \theta^{t+1} & = & F(\theta_1^t, \theta_2^t, \ldots, \theta_K^{t})
\label{eqn:advfed}
\end{eqnarray}



The general optimization framework for solving Eqn.\eqref{eqn:advfed} is given in Algorithm~\ref{algo:FedAT}.
We consider two specific implementations of the fusion algorithm: Federated Averaging Adversarial Training (\texttt{FedAvgAT}) and Federated Curvature Adversarial Training (\texttt{FedCurvAT}). These fusion algorithms differ in the choice of their loss function $\Lagr$ and fusion operator $F$. \texttt{FedAvgAT} uses the cross-entropy loss function and weighted average of model weights as fusion operation. \texttt{FedCurvAT} in turn employs regularized loss function and an extended fusion involving the model weights and some auxiliary parameters. 
\begin{algorithm}[]
\small
\caption{\small \textbf{Federated adversarial training framework}}
 \label{Federated_adversarial_training_algo}
 \begin{algorithmic}
 \STATE {\bfseries Input:} number of clients: $K$, number of communication rounds: $R$, PGD Attack $A_{T, \epsilon, \alpha}$ with 3 parameters: $T, \epsilon, \alpha$(number of PGD steps, perturbation ball size, step size) , number of local training epochs per round: $E$, Initial model weights: $\theta^{0}$
 
 \FOR{$t=0$ {\bfseries to} $R-1$}
 \STATE Server sends $\theta^{t}$ to all devices \\
 \FOR{client $k=0$ {\bfseries to} $K-1$}
  \STATE $\theta^{t}_{k}$ = \at(E, $\theta^{t}$, $A_{T, \epsilon, \alpha}$) \\
  \STATE Send $\theta^{t}_{k}$ back to server \\
 \ENDFOR
 \STATE \COMMENT {Server performs aggregation over agent updates} \\ $\theta^{t+1} = F(\theta_1^t, \theta_2^t, \ldots, \theta_K^{t})$ \\
 \ENDFOR
  \end{algorithmic}
  \label{algo:FedAT}
\end{algorithm}

\subsection{\fedavgat Algorithm} 
The first fusion algorithm we consider is Federated Averaging \cite{mcmahan2016communicationefficient} which is a widely used algorithm in federated learning.



At the end of round $t$, each client sends its model parameters, $\theta^{t}_{k}$, back to server. The server performs a weighted average of the model parameters to obtain the aggregate model parameters $\theta^{t+1}$ for the next round as:
\begin{equation}
    \label{eqn:fedavgupdate}
    \theta^{t+1} = \frac{1}{|D|}\sum_{k}|D_k| \theta^{t}_{k}
\end{equation}

\subsection{\fedcurvat Algorithm} 
\label{fed_curvature}
With non-iid data distribution between clients, local models tend to drift apart and inhibit learning using \fedavg. In \fedcurv~\cite{shoham2019overcoming}, a penalty term is added to the local training loss, compelling all local models to converge to a shared optimum.

At any client $k$, the loss function used for training $\Lagr(f_{\theta}(\Tilde{x}_{k}), y_{k})$ comprises of two components: 

\noindent{1.} Training loss on local data, $L_{k}(\theta)$.

\noindent{2.} Regularization loss, $R_k(\theta)$, given by:
  \[R_k(\theta) = \small \sum\limits_{j \in K \setminus k} (\theta - \hat{\theta}_{t-1,j}) \text{diag}(\hat{\mathcal{I}}_{t-1,j}) (\theta - \hat{\theta}_{t-1,j})\].

At the beginning of round $t$, starting from the initial point $\theta^t$ given by Eqn.(\ref{eqn:fedavgupdate}), each node optimizes $L_{k}(\theta) + \lambda R_k(\theta)$. 
At end of round $t$, client $j$ sends its local model weights $\hat{\theta}_{t,j}$ and the diagonal of Fisher Information Matrix $\hat{\mathcal{I}}_{t,j}$, evaluated on its local dataset and local model weights. 
$\lambda$ is the regularization hyperparameter which balances the tradeoff between local training, and model drift from the local client weights at the end of previous round.

\section{Impact of adversarial training on federated model performance}
\label{sec:challenges}
In this section, we empirically outline the primary challenges in adopting \at to federated setting. These challenges provide useful insights and motivate our proposed solution. We use \fedavg as our baseline fusion algorithm and compare the relative impact on performance due to \fedavgat.  

\subsection*{Datasets and model architecture for evaluation}
\label{subsec:data}
We conduct our evaluation with two datasets: CIFAR-10~\cite{cifar10} and EMNIST-Balanced~\cite{cohen2017emnist}. We use Network-in-Network (NIN)~\cite{lin2014network} and VGG-9 network~\cite{simonyan15} for training on the CIFAR-10 dataset.
For experiments with EMNIST-Balanced, we use a network with two convolution layers followed by two fully connected layers. We refer to this model as EMNIST-M in subsequent sections.

\subsection{Increased drop in natural and adversarial accuracy with federated \at and non-iid data}
\label{subsec:non-iid}
In federated learning deployments, data partitions across clients invariably exhibit non-iid behavior \cite{mcmahan2016communicationefficient}. Thus, we begin by investigating the effect of adversarial training with non-iid data. 
Details for the data partition strategy for non-iid data can be found in Section.~\ref{non-iid split} 
Table~\ref{Adversarial_challenge} reports the performance of \fedavgat for both natural training and adversarial training for all the three models. The number of local training epochs, $E=1$. We also report results for centralized training (entire dataset available at the training site) for better comparison.

First, we observe that adversarial training with iid data distribution leads to performance that is close to that of centralized setting for all the three models. Second, as has been indicated in prior work~\cite{shoham2019overcoming}, we validate that using \fedavg with non-iid data, leads to a drop in accuracy even for natural training. However, as the main focus of our paper, we note a much larger drop in both natural and adversarial accuracies when performing adversarial training on non-iid data when compared to iid data. It is most severe for EMNIST-M model, where the adversarial accuracy drops from 72\% in the centralized setting to 58\%.


\begin{table}[!ht]
\small
\centering
\setlength{\tabcolsep}{3pt}
\begin{tabular}{p{1pt}ccccc}
\multicolumn{5}{c}{} \\
\cline{2-6}
 &  & & NIN & VGG & EMNIST-M \\
 \cline{2-6}
& Natural & Centralized & (88\%, 0\%) & (91\%, 0\%) & (87\%, 0\%) \\ [-0.05in]
&& iid & (87\%, 0\%) & (90\%, 0\%) & (86.5\%, 0\%)
 \\ [-0.05in]
&& non-iid & (78.5\%, 0\%) & (86\%, 0\%) & (82.5\%, 0\%) \\ \cline{2-6}
& Adversarial & Centralized & (75\%, 42\%) & (75\%, 44\%) & (84.5\%, 72\%) \\ [-0.05in]
&& iid & (74.5\%, 42\%) & (75, 44\%) & (84\%, 72\%)\\ [-0.05in]
&& non-iid & (67\%, 36\%) & (68.5\%, 36\%) & (84\%, 58\%)\\ 
\cline{2-6}
\end{tabular}
\caption{\small Performance for natural and adversarial training for centralized and \fedavgat with K=5 clients. The performance is reported as a tuple: (natural accuracy, adversarial accuracy).}
\label{Adversarial_challenge}
\end{table}

We hypothesize increased model drift with non-iid data as the reason for the significant reduction in model performance with \fedavgat. Model drift refers to models learning very different representations of the data. At the start of a round, every client begins local adversarial training with the same model weights. However, the generated adversarial examples depend on both the local samples and the local model. As the local training progresses, the adversarial examples at the clients can become highly tuned towards the local data and model and may differ a lot from adversarial examples generated at other clients. Hence the models which are trained on these samples can exhibit significant drift. We validate our hypothesis and evaluate model drift in detail using SVCCA~\cite{raghu:svcca}  in Section~\ref{subsec:svcca}.

\subsection{Increased communication overhead} \label{sec:communication_overhead_section}
\begin{figure}[t]
\centering
\includegraphics[scale=0.5]{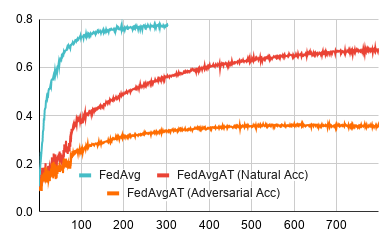} 
    \caption{\centering \small Accuracy vs communication round plot for \fedavg (natural training) and \fedavgat with 5 clients and non-iid data distribution using NIN.}
    \label{fig:communication_challenge}
\end{figure}

In this section, we study the effect of adversarial training on the convergence rate of the global model.
The convergence plots for \fedavg and \fedavgat on non-iid data between $K=5$ clients, with $E=1$ are shown in Fig.~\ref{fig:communication_challenge}. While \fedavg converges within $300$ communication rounds, \fedavgat takes about $800$ rounds to converge. This highlights that adversarial training requires more communication rounds compared to natural training in a federated setup. With faster compute infrastructure at the clients, increased communication cost is often the primary bottleneck for training of the models in a federated setting. 




\section{\feddynat Algorithm}
To address the twin challenges described in Section \ref{sec:challenges}, we propose a new algorithm, Federated Dynamic Adversarial Training (\feddynat). While prior work~\cite{Wang2019AdaptiveCS, spiridonoff2020local} have studied the tradeoff between communication overhead, and the convergence accuracy for natural training with iid data distribution. In this paper, we present this trade-off for adversarial training with non-iid data distribution.

\subsection{Influence of \mbox{$E$} on convergence} \label{influenceE_convergence}
To motivate \feddynat, we first discuss the \emph{sensitivity} of \at in a federated setting with respect to the number of local adversarial training epochs, $E$, between successive model aggregation. 
We vary the value of $E$ and study its impact on convergence speed as well as accuracy achieved.

\begin{figure}[!ht]
\centering
\includegraphics[scale=0.5]{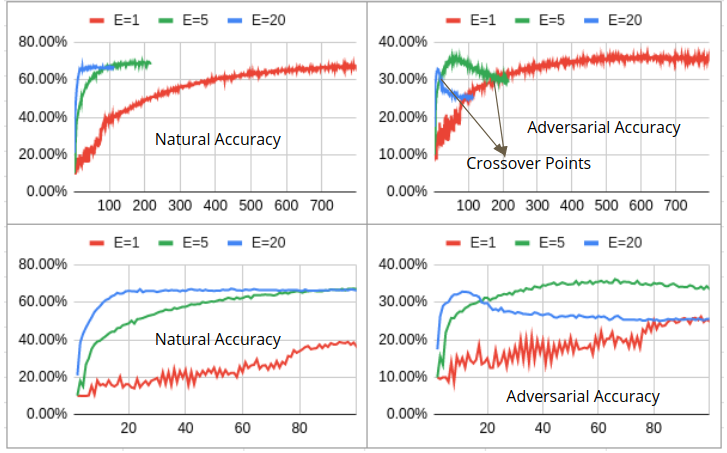}
    \caption{\centering \small The top row shows the accuracy vs.\ communication round plots with \fedavgat for $E=1, 20, 50$ on NIN. The bottom row is a zoomed version of the convergence plot, for first 100 communication rounds.}
    \label{fig:E_switch.png}
\end{figure}

The convergence plots for \fedavgat for $E = 1, 5$ and $20$ are presented in Fig.~\ref{fig:E_switch.png}. Setting $E=1$ leads to the best performance (for both natural and adversarial accuracy) but takes $800$ communication rounds to converge. The adversarial accuracy plots for $E=5$ and $E= 20$ converge much faster to almost the same accuracy as $E=1$ (after $800$ rounds), however, as training progresses the accuracy drops significantly. 

A higher value of $E$ is found to benefit the convergence speed for natural training as shown in \cite{mcmahan2016communicationefficient}. However, the fall in adversarial accuracy  with higher $E$ in the later phase of training can be explained with the findings in earlier studies \cite{pmlr-v97-wang19i,gupta20}. They show that the adversaries generated only in the later phase of the  training are  potent to the final model. A higher value of $E$ results in models drifting too much from each other due to non-iid data distribution (and hence the locally generated adversaries) but it starts to impact only in the later phase of the training. 




\subsection{\fedcurvat improves performance over \fedavgat with higher $E$} \label{FedCurvature_improves_performance_highE_fedavg}


\fedcurv, as described in  Section.~\ref{fed_curvature}, has been shown to result in better performance as compared to \fedavg in the context of non-iid data distribution~\cite{shoham2019overcoming}. We evaluate performance of \fedcurvat on non-iid data and its dependency on $E$. 
Table~\ref{FedAvg_FedCurvature_non-iid_highE} compares the performance of \fedcurvat and \fedavgat for different values of $E$ for NIN, VGG and EMNIST-M models. 

We see a significant improvement in adversarial accuracy with \fedcurvat compared to \fedavgat. Further, the improvement increases as we increase $E$. We explain the dependence of model drift on $E$ in Section~\ref{subsec:svcca}. We attribute this improvement to \fedcurvat's regularized loss which helps in reducing model drift. 
Increased drift of local models affects the adversarial accuracy of the global model, reflected in \fedavgat results for $E=20$ and $E=50$.
\begin{table}[!hbt]
\centering
\small
\setlength{\tabcolsep}{3pt} 
\begin{tabular}{p{1pt}cccrr}
\multicolumn{5}{c}{} \\
\cline{2-6}
 & Model & E & K & \fedavgat & \fedcurvat\\ 
 \cline{2-6}
& NIN & 20 & 2 & (62.1\%, 27.5\%) & (\textbf{66.6}\%, \textbf{33.5}\%)\\ [-0.05in]
&&& 5 & (67.1\%, 27.8\%) & (\textbf{66.6}\%, \textbf{31}\%) \\ [-0.05in]
&&& 10 & (61.6\%, 25.4\%) & (\textbf{62.4}\%, \textbf{30.1}\%) \\
 \cline{3-6}
&& 50 & 5 & (63.8\%, 24.5\%) & (\textbf{65.2}\%, \textbf{32}\%) \\ [-0.05in]
&&& 10 & (60.5\%, 21.7\%) & (\textbf{60.9}\%, \textbf{30.7}\%) \\
 \cline{2-6}
\end{tabular}
\quad
\begin{tabular}{p{1pt}cccrr}
\multicolumn{5}{c}{} \\
\cline{2-6}
 & Model & E & K & \fedavgat & \fedcurvat\\ 
 \cline{2-6}
& VGG & 20 & 2 & (60.2\%, 26.7\%) & (\textbf{67.1}\%, \textbf{30.5}\%)\\ [-0.05in]
&&& 5 & (67.7\%, 31.3\%) & (\textbf{68.3}\%, 31.3\%) \\ [-0.05in]
&&& 10 & (65.1\%, \textbf{29}\%) & (\textbf{65.6}\%, 28.3\%) \\
 \cline{3-6}
&& 50 & 5 & (64\%, 30.6\%) & (\textbf{66.8}\%, \textbf{32}\%) \\ [-0.05in]
&&& 10 & (\textbf{64}\%, 28.5\%) & (\textbf{64}\%, 28.7\%) \\
\cline{3-6}
\end{tabular}
\quad
\begin{tabular}{p{1pt}cccrr}
\multicolumn{5}{c}{} \\
\cline{2-6}
 & Model & E & K & \fedavgat & \fedcurvat\\ 
 \cline{2-6}
& EMNIST-M & 5 & 2 & (83.5\%, \textbf{60.5}\%) & (\textbf{84.5}\%, 60\%) \\ [-0.05in]
 &&& 5 & (80\%, 53.3\%) & (\textbf{81}\%, \textbf{55}\%) \\ [-0.05in]
 &&& 10 & (78.8\%, 48.4\%) & (\textbf{79.9}\%, \textbf{50}\%) \\
 \cline{3-6}
&& 20 & 2 & (80.3\%, 52.6\%) & (\textbf{83.1}\%, \textbf{56}\%)\\ [-0.05in]
&&& 5 & (74.1\%, 47\%) & (\textbf{78.6}\%, \textbf{50.4}\%)\\ [-0.05in]
&&& 10 & (72.8\%, 42.5\%) & (\textbf{76}\%, \textbf{45.7}\%) \\ 
 \cline{2-6}
\end{tabular}

\caption{\centering \small \fedavgat vs \fedcurvat for various $E$ values after 50 communication rounds, for NIN, VGG and EMNIST-M models.}
\label{FedAvg_FedCurvature_non-iid_highE}
\end{table}

\subsection{Our algorithm : \feddynat}
\label{subsec:feddnyat}

The experiments (in Sections.~\ref{influenceE_convergence} and \ref{FedCurvature_improves_performance_highE_fedavg}) lead to two primary observations: (i) A high value of $E$ leads to large model drift affecting the adversarial accuracy of the global model. A low value of $E$ results in a higher adversarial accuracy at convergence but converges slowly, resulting in high communication cost. (ii) \fedcurvat outperforms \fedavgat for high values of $E$, and the improvement increases with $E$.

We combine the insights in our algorithm \feddynat, which follows a dynamic schedule for the number of local training epochs at each round. The training starts with a high value of $E = E_{0}$, and decays its value every $F_{E}$ rounds by a decay factor of $\gamma_{E}$. Further, we use \fedcurv as the fusion algorithm (but it does not preclude the use of \fedavg for fusion). Details of the algorithm are outlined in Algorithm ~\ref{FedDynAT_algo}.

\begin{algorithm}[]
\small
\caption{\small \textbf{\feddynat}}
 \label{FedDynAT_algo}
 \begin{algorithmic}
 \STATE {\bfseries Input:} number of clients: $K$, number of communication rounds: $R$, PGD Attack: $A_{T, \epsilon, \alpha}$, Initial model weights: $\theta^{0}$, Local training epochs at start: $E_{0}$, Decay rate: $\gamma_{E}$, 
 Decay frequency: $F_{E}$
 
 \FOR{$t=0$ {\bfseries to} $R-1$}
 \STATE $E_{t}= \ceil*{E_{0}*\gamma_{E}^{\floor*{\frac{t}{F_{E}}}}} $ \\
 \STATE Server sends $\theta^{t}$ to all devices \\
 \FOR{client $k=0$ {\bfseries to} $K-1$}
  \STATE $\theta^{t}_{k}$ = \at($E_t$, $\theta^{t}$, $A_{T, \epsilon, \alpha}$) \\
  \STATE Send $\theta^{t}_{k}$ back to server \\
 \ENDFOR
 \STATE \COMMENT {Server performs aggregation over agent updates} \\ $\theta^{t+1} = \fedcurv(\theta_1^{t}, \ldots, \theta_{K}^{t})$ \\
 \ENDFOR
  \end{algorithmic}
\end{algorithm}

\section{Experiments and Results}

\subsection{Experimental set-up}
We list here the implementation details, and the non-iid data partition strategy across clients. All experiments were conducted on a VM with eight Intel(R) Xeon(R) CPU E5-2690 v4 @ 2.60GHz CPUs and one NVIDIA V100 GPU. 

\subsubsection{non-iid data split} \label{non-iid split}
We conduct all our experiments with $K$ clients, $K \in \{2, 5, 10\}$. 
To achieve non-iid data distribution, we introduce skew in the data split. 
Each client gets data corresponding to all the classes, but a majority of the data samples are from a subset of classes. We denote the set of majority classes at client $k$ as $M_{k}$ with ${\cal M}$ being the set of all classes in the dataset (e.g., for CIFAR-10 with 10 classes $|{\cal M}|=10$). ${\cal M}\setminus M_k$ will be the set of minority classes. We keep  $M_{k}$ as mutually exclusive across the clients.
We ensure that each client has roughly the same number of data samples.

We characterize the data split with a skew parameter $s$ which is a percentage of the aggregate data size corresponding to a class in ${\cal M}$. We first create $\{M_k\}$ by dividing all class labels in ${\cal M}$ equally among $K$ clients, i.e.$|M_k|=|{\cal M}|/K$. We then divide the total data across clients such that a client has $s\%$ of data for any class in ${\cal M} \setminus M_k$ and $(100-(K-1)*s)\%$ of data for a class in $M_k$. 
Table~\ref{skew_values} shows the skew values we have used for our non-iid data partitions.

\begin{table}[!ht]
\centering
\small
\setlength{\tabcolsep}{6pt} 
\begin{tabular}{p{1pt}cccc}
\multicolumn{4}{c}{} \\
\cline{2-5}
 & Dataset & $K=2$ & $K=5$ & $K=10$ \\
 \cline{2-5}
& CIFAR-10 & 1\% & 2\% & 2\% \\
& EMNIST & 0.1\% & 0.1\% & 0.1\% \\
\cline{2-5}
\end{tabular}
\caption{\centering \small Skew parameter $s$ value used in our experiments.}
\label{skew_values}
\end{table}


\subsubsection{Implementation details}
All our models (NIN, VGG-9, and EMNIST-M) are implemented in Keras~\cite{chollet2015keras}. 
We use the PGD adversarial attack $A_{T, \epsilon, \alpha}$ which has 3 parameters: $T, \epsilon, \alpha$ (number of steps, ball size, step size). For CIFAR-10, we use $T=10, \epsilon= \frac{8}{255}, \alpha=\frac{2}{255}$. For EMNIST, we use  $T=10, \epsilon= \frac{3}{10}, \alpha=\frac{1}{20}$.
\at at each client is done with a batch size of $32$. For CIFAR-10, we use the SGD optimizer with a constant learning rate $=10^{-2}$ and momentum $=0.9$. For EMNIST, we use the Adam optimizer with a constant learning rate $=10^{-4}$.

To train the VGG network, we use data augmentation (random horizontal/vertical shift by 0.1 and horizontal flip). We do not use data augmentation for NIN and EMNIST-M networks.
We ignore all batch normalization \cite{10.5555/3045118.3045167} layers in the VGG architecture.


\subsection{\feddynat results}
We compare \feddynat to \fedavgat and \fedcurvat baselines, both of which are run with a fixed value of $E$. For a given communication budget, we evaluate \fedavgat and \fedcurvat with various values of $E$ and report performance with $E$ resulting in best natural accuracy for the global model.
We report two observations in this section
\begin{itemize} [noitemsep,nolistsep]
    \item \feddynat outperforms \fedavgat, \fedcurvat for a fixed communication budget.
    \item \feddynat achieves similar performance as \fedavgat(E=1), 
    albeit in significantly lower number of communication rounds.
\end{itemize}

\subsubsection*{Results with varying $E$-schedule}
\begin{figure}[!ht]
\centering
\includegraphics[scale=0.5]{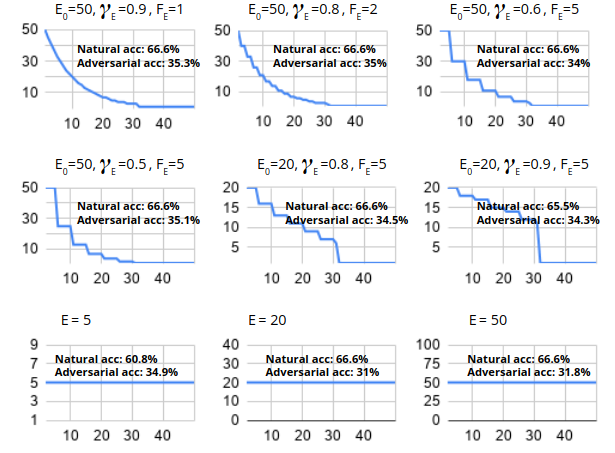} 
    \caption{\centering \small Performance of dynamic $E$ schedule with various parameter configurations. x-axis shows communication rounds and y-axis is the value of $E$ used in that round.}
    \label{fig:E_schedule_FedDynAT}
\end{figure}

The dynamic $E$-schedule is defined by 3 parameters: the initial value $E_{0}$, decay rate $\gamma_{E}$ and decay frequency $F_{E}$ as described in Section~\ref{subsec:feddnyat}. We report the results for various combination of these parameters in  
Fig.~\ref{fig:E_schedule_FedDynAT}. The bottom row shows \fedcurvat results for 3 different fixed values of $E$. The top rows show the results for various configurations of the dynamic $E$ schedule. We notice an improved performance in both natural and adversarial accuracy with \feddynat with a smooth drop in $E$ compared to \fedcurvat with fixed $E$. 


\subsubsection{Accuracy with different communication budgets}
We set out to solve the adversarial training problem with non-iid data split in a communication constrained setting. We vary the communication buget i.e maximum number of communication rounds and demonstrate how \feddynat performs as compared to \fedavgat and \fedcurvat. We evaluate \fedavgat and \fedcurvat with multiple values of $E$ and report performance corresponding to the best natural accuracy of the global model.  Table~\ref{FedDynAT_NIN_100rounds} shows the results for Network-in-network model for CIFAR10 dataset and  Table~\ref{FedDynAT_EMNIST_20rounds} shows that of EMNIST-M model for EMNIST dataset. We defer results with VGG to the Supplement.

\begin{figure}[!ht]
\centering
\includegraphics[scale=0.5]{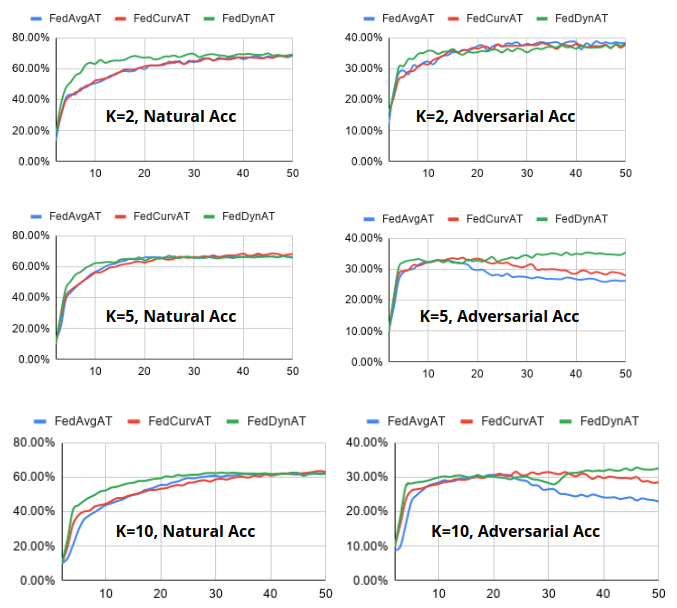} 
    \caption{\centering \small Accuracy vs communication round for \fedavgat, \fedcurvat and \feddynat with NIN for a communication budget of 50 rounds.}
    \label{fig:convergence_plot_50rounds}
\end{figure}

The convergence plot for \fedavgat, \fedcurvat and \feddynat with NIN with a budget of 50 rounds is presented in Fig.~\ref{fig:convergence_plot_50rounds}. Notice how the natural accuracy curves for all 3 algorthms converge to similar numbers but the adversarial accuracy with \feddynat is significantly better than \fedavgat and \fedcurvat. 

\noindent{\bf FedAvg fusion with dynamic $E$: } 
\feddynat uses the FedCurvature fusion algorithm. We consider another baseline where we replace the fusion algorithm with FedAvg and refer to it as \fedavgdynat.  
The results with NIN in Table~\ref{FedDynAT_NIN_100rounds} demonstrate that \feddynat  outperforms this baseline. In Section~\ref{FedCurvature_improves_performance_highE_fedavg} we established that \fedcurvat results in improvement over \fedavgat for higher $E$. A high $E_{0}$ for our E-schedule contributes to better eventual performance with \feddynat over \fedavgdynat.
We find \fedavgdynat performs similar to \feddynat for EMNIST dataset for the particular communication budget. 

\begin{table}[!ht]
\centering
\small
\setlength{\tabcolsep}{2pt} 
\begin{tabular}{p{1pt}ccrrrr}
\multicolumn{6}{c}{} \\
\cline{2-7}
 & K & Budget & \fedavgat & \fedcurvat & \fedavgdynat & \feddynat \\
 \cline{2-7}
& 2 & 50 & (68.2\%, \textbf{38.5}\%) & (68.7\%, 37.8\%) & (67.2\%, 34.3\%) & (\textbf{68.4}\%, 38\%) \\ [-0.05in]
&& 100 & (70.0\%, 38\%) & (\textbf{71}\%, 37.5\%) & (67.2\%, 35.3\%) & (70.4\%, \textbf{38}\%) \\
\cline{2-7}
& 5 & 50 & (\textbf{67.1}\%, 27.8\%) & (66.6\%, 31\%) & (66.5\%, 34.8\%) & (66.6\%, \textbf{35.3}\%) \\ [-0.05in]
&& 100 & (64.9\%, 35\%) & (65.8\%, 35.2\%) & (66.8\%, 35\%) & (\textbf{66.8}\%, \textbf{35.9}\%) \\
\cline{2-7}
& 10 & 50 & (61.6\%, 25.4\%) & (61.9\%, 30.7\%) & (61.8\%, 30\%) & (\textbf{62.3}\%, \textbf{32.5}\%) \\ [-0.05in]
&& 100 & (61.6\%, 25.4\%) & (61.9\%, 30.7\%) & (61.9\%, 31.4\%) & (\textbf{62.6}\%, \textbf{33.3}\%) \\
 \cline{2-7}
\end{tabular}
\caption{\centering \small \feddynat performance on Network-in-network model for communication budgets of 50 and 100 rounds. We evaluate \fedavgat and \fedcurvat with $E=\{1, 5, 20\}$ and report the best individual performance.}
\label{FedDynAT_NIN_100rounds}
\end{table}

\begin{table}[!ht]
\centering
\small
\setlength{\tabcolsep}{2pt} 
\begin{tabular}{p{1pt}ccrrrr}
\multicolumn{6}{c}{} \\
\cline{2-7}
 & K & Budget &  \fedavgat & \fedcurvat & \fedavgdynat & \feddynat \\
 \cline{2-7}
& 2 & 20 & (83.5\%, 57.5\%) & (83.8\%, 60\%) & (84\%, 64\%) & (\textbf{84.5}\%, \textbf{64.2}\%) \\ [-0.05in]
&& 50 & (84.1\%, 65.5\%) & (84.1\%, 65.5\%) & (84\%, 64\%) &(\textbf{84.5}\%, \textbf{64.5}\%) \\
\cline{2-7}
& 5 & 20 & (78.4\%, 50.6\%) & (81\%, 55\%) & (82.1\%, \textbf{56.5}\%) & (\textbf{82.3}\%, 56.2\%) \\ [-0.05in]
&& 50 & (80.7\%, 54.3\%) & (81\%, 55\%) & (83.4\%, \textbf{59}\%) & (\textbf{83.6}\%, 58.3\%) \\
\cline{2-7}
& 10 & 20 & (75.8\%, 45.4\%) & (76.8\%, 45.8\%) & (78.9\%, \textbf{51.5}\%) & (\textbf{79.9}\%, 51.3\%) \\ [-0.05in]
&& 50 & (78.3\%, 47.5\%) & (79.9\%, 50\%) & (\textbf{81.9}\%, \textbf{54.8}\%) & (81.7\%, 54.7\%) \\
\cline{2-7}
\end{tabular}
\caption{\centering \small \feddynat performance on EMNIST-M compared to \fedavgat, \fedcurvat with a communication budget of 20 and 50 rounds. We evaluate \fedavgat and \fedcurvat with $E=\{1, 5, 20\}$ and report performance with $E$ resulting in best natural accuracy.}
\label{FedDynAT_EMNIST_20rounds}
\end{table}

\subsubsection{Convergence speed for \feddynat} \label{convergence_speed_section}
\begin{figure}[]
\includegraphics[width=1.0\columnwidth]{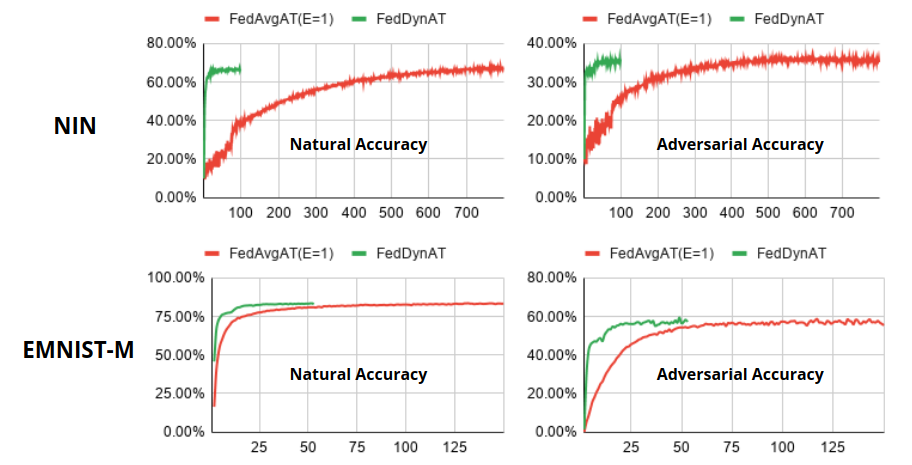} 
    \caption{\small Accuracy vs communication round plot for \feddynat and \fedavgat evaluated with E=1, on $K=5$ client non-iid data split with NIN and EMNIST-M.} 
    \label{fig:convergence_plot_fedavg_E=1_FedDynAT}
\end{figure}

Figure~\ref{fig:convergence_plot_fedavg_E=1_FedDynAT} shows that convergence is much faster with \feddynat compared to \fedavgat(E=1) for both NIN and EMNIST-M. 
For reaching the same accuracy, with NIN, \feddynat converges in 100 communication rounds while \fedavgat takes 800 rounds to converge. Similarly, 
with EMNIST-M, \feddynat converges in 50 communication rounds while \fedavgat takes 150 rounds to converge.
Results for other values of $K$ are included in the appendix.

\subsection{Understanding performance through model drift: A SVCCA-based Analysis }
\label{subsec:svcca}


 Singular Vector Canonical Correlation Analysis (SVCCA) \cite{raghu:svcca} is a technique to compare two network representations that is invariant to affine transforms. SVCCA interprets each neuron as a vector (formed by considering the neuron activations on the input data points) and a layer as subspace spanned by that layer's neurons. 
 
 We measure the SVCCA scores (in range [0-1]) between the corresponding layer of two models. A high score implies similar representations of two layers (or lower drift), and low score implies that the two layers have learnt different representations (indicating a higher drift). We show results for (i) dependence of model drift on $E$ (ii) comparison of model drift with and without \at in federated learning; and (iii) improvement in model drift due to \feddynat over \fedcurvat and \fedavgat.

We report results for two of the five client NIN models\footnote{NIN model: x $\rightarrow$ conv1 $\rightarrow \ldots \rightarrow$ conv9 $\rightarrow$ fc $\rightarrow$ y}. The plots show SVCCA scores for convolution layers $5-8$ where the drift is more pronounced. Layers $1-4$, act as feature extractor layers and show relatively low drift.

\begin{figure}[t]
\centering
\includegraphics[scale=0.5]{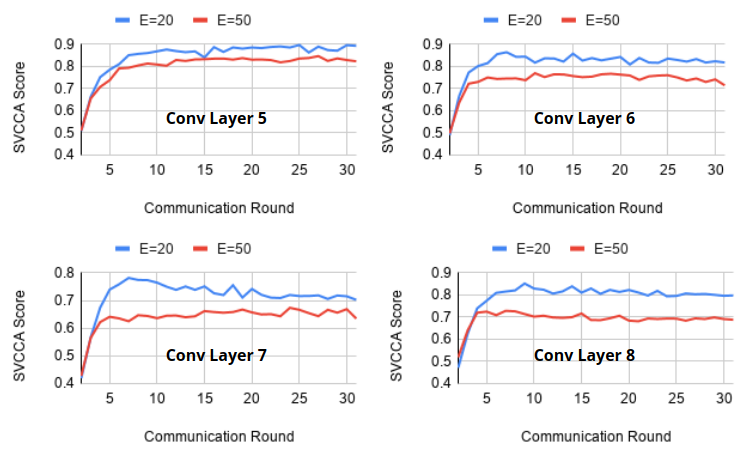} 
    \caption{\centering \small SVCCA scores evaluated on 2 out of 5 clients for \fedavgat with $E=20$ and $E=50$ using NIN.}
    \label{fig:svcca_E_20_50}
\end{figure}

\noindent{\bf Increased model drift with higher E:} With a higher value of $E$, local models train more on local data, consequently the model drift is higher. We illustrate the model drift phenomenon in Fig.~\ref{fig:svcca_E_20_50}, by running \fedavgat with $E=20$ and $E=50$. The SVCCA scores with $E=20$ are much higher than those for $E=50$, indicating a lower model drift for $E=20$.

\begin{figure}[!ht]
\centering
\includegraphics[scale=0.5]{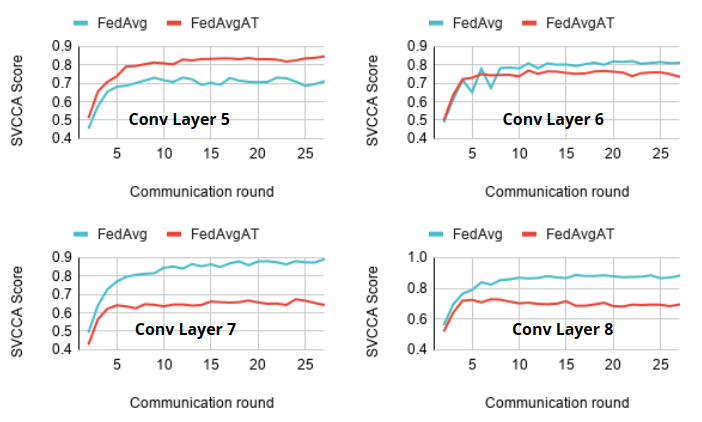}
    \caption{\centering \small SVCCA scores with $E=50$ for both adversarial and natural training with NIN.}
    \label{fig:svcca_natural_adversarial.png}
\end{figure}

\noindent{\bf Comparing model drift with \fedavg and \fedavgat on non-iid data:} In Fig.~\ref{fig:svcca_natural_adversarial.png}, we compute SVCCA scores for \fedavg(natural training) and \fedavgat with $E=50$. We observe that the similarity scores for \fedavgat are generally lower than \fedavg indicating higher drift between the layers. This shows that the drift is higher for adversarial training. 
Furthermore, this drift is more prominent in the higher layers of the model (close to the output). 

\begin{figure}[!ht]
\centering
\includegraphics[scale=0.5]{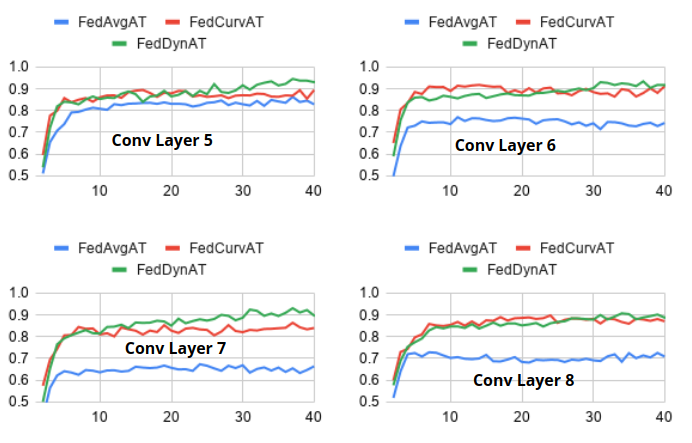}
    \caption{\centering \small SVCCA scores evaluated on 2 out of 5 clients for \fedavgat, \fedcurvat and \feddynat with NIN. \fedcurvat and \fedavgat use a constant $E=50$ for local training. \feddynat uses $E_{0}=50, \gamma=0.5, F_{E}=5$ for the dynamic E-schedule.}
    \label{fig:svcca_fedavg_fedcurvature}
\end{figure}

\noindent{\bf Comparing model drift between \feddynat, \fedavgat, \fedcurvat for non-iid data:} In Fig.~\ref{fig:svcca_fedavg_fedcurvature}, we present a comparison of SVCCA scores for \fedavgat, \fedcurvat and \feddynat, all the three techniques we used for adversarial training.
We observe that the similarity score for \fedcurvat is significantly higher than \fedavgat, which validates the proposition that \fedcurvat arrests the local model drift. 

However, \feddynat is able to beat both \fedcurvat and \fedavgat in SVCCA scores. use a constant $E=50$ for local training. \feddynat uses $E_{0}=50, \gamma=0.5, F_{E}=5$ for the E-schedule. As training progresses, and the value of $E$ decreases, less epochs of local model training contributes to a lower model drift. The layers closer to the input have not been shown in the plots as they show low model drift for all 3 algorithms.

\section{Related Work}
To address communication overhead in distributed learning, algorithms perform multiple local epochs of training before synchronization and thus reduce the communication overhead~\cite{konecny16,mcmahan2016communicationefficient,NEURIPS2019_4aadd661}. The convergence analysis of these algorithms connects the rate of convergence with number of clients and local training epochs~\cite{yu2019linear,yu2018parallel}. These results are mostly  obtained under strong assumptions on the degree of dissimilarity (i.e., non-iidness) in the data distribution between clients which may not hold in practice. Of particular relevance to our work is~\cite{Wang2019AdaptiveCS} which analyses tradeoff between decreased wall clock time for convergence and increase in convergence error with increasing $E$. Based on the analysis for iid distribution, an adaptive algorithm is proposed that dynamically changes the number of local training rounds. 

Several algorithms have been proposed to achieve faster convergence while reducing communication overhead with non-iid data. ~\cite{li2018federated, shoham2019overcoming} use additional regularization terms in the client loss function to minimize divergence between local and global models. 
Going beyond simple averaging of weights during fusion, ~\cite{pmlr-v97-yurochkin19a, wang2020federated} perform layer-wise matching of neurons from different clients, before averaging the model weights leading to improved communication efficiency. To prevent interference between incompatible tasks across clients, ~\cite{yoon2020federated} selectively transfers the knowledge between clients. All the above fusion algorithms have been studied for natural training only.

In centralized setting \at by ~\cite{madry2018towards} is shown to provide some of the strongest defense against PGD attacks~\cite{pmlr-v80-athalye18a} and is a widely accepted benchmark~\cite{hendrycks2019pretraining, pmlr-v80-athalye18a}. In concurrent work,~\cite{zizzo2020fat} has also implemented traditional \at in federated setting. Different from our work, their focus was on evaluating the vulnerability of both Byzantine resilient defenses and federated \at, rather than mitigating the performance impact of \at in federated setting.

\section{Conclusion}
We proposed a novel algorithm for performing adversarial training of deep neural networks in a federated setting. We showed that federated setting specially with non-iid data distribution significantly affects the training performance due to model drift and takes much longer to converge as compared to natural training. We showed that the proposed algorithm is able to address the above challenges effectively and is able to get comparable or better accuracy in significantly lower number of rounds. 
Incorporating dynamic scheduling with more advanced model fusion techniques remains a topic of future research. 

\newpage
\bibliographystyle{abbrv}
\bibliography{icml}
\clearpage
\section{Supplementary material}

\subsection{Tuning hyperparameter lambda for \fedcurvat}
$\lambda$ is the regularization hyperparameter in the loss function for \fedcurvat. It balances the tradeoff between local training, and model drift from the local client weights at the end of previous round. A very high value of $\lambda$ inhibits local training while a very low value of $\lambda$ is akin to \fedavgat.

We do a gridsearch to tune $\lambda$ over a range varying from $\lambda=0$ upto $\lambda=100$. 
For $K=2$ clients non-iid data, with $E=20$, we use $\lambda = 5$ for NIN, $\lambda=5$ for VGG, and $\lambda = 1e^{-2}$ for EMNIST-M.

From Section~\ref{subsec:svcca}, we observed that there is an increased model drift with a higher $E$. 
To compensate for the drift, we increase $\lambda$ on increasing $E$. 

Similarly, we use a higher value of $\lambda$ as the number of clients $K$ increases. 
As mentioned in Section~\ref{non-iid split}, our approach to generate non-iid distribution for a given dataset results in a decrease in the number of majority classes per client, on increasing $K$. This increases
the degree of non-iidness in the data distribution on increasing $K$, which explains the need for higher $\lambda$.  

\subsection{Additional results}

\subsubsection{\feddynat performance with VGG network}

In Table~\ref{FedDynAT_VGG_100rounds}, we vary the communication budget and demonstrate the improvement \feddynat offers compared to \fedavgat and \fedcurvat for a communication budget of 20 rounds as well as 50 rounds. 

\begin{table}[!ht]
\centering
\small
\setlength{\tabcolsep}{2pt} 
\begin{tabular}{p{1pt}ccrrrr}
\multicolumn{6}{c}{} \\
\cline{2-7}
 & K & Budget & \fedavgat & \fedcurvat & \fedavgdynat & \feddynat \\
 \cline{2-7}
& 2 & 20 & (67.6\%, \textbf{39.3}\%) & (67.6\%, \textbf{39.3}\%) & (67.3\%, 35.5\%) & (\textbf{68}\%, 38\%) \\ [-0.05in]
&& 50 & (70.1\%, 35.4\%) & (70.1\%, 35.4\%) & (69.3\%, 35.9\%) & (\textbf{70.4\%}, \textbf{39\%}) \\
\cline{2-7}
& 5 & 20 & (67.3\%, 31\%) & (67.9\%, 31.3\%) & (67.5\%, 35.7\%) & (\textbf{68.5\%}, \textbf{35.8\%}) \\ [-0.05in]
&& 50 & (68.5\%, 35\%) & (68.5\%, 35\%) & (\textbf{69.1\%}, 35\%) & (\textbf{69.1\%}, \textbf{35.6\%}) \\
\cline{2-7}
& 10 & 20 & (64\%, 30.6\%) & (64\%, 29.7\%) & (\textbf{64.3\%}, 32.2\%) & (\textbf{64.3\%}, \textbf{32.3\%})  \\ [-0.05in]
&& 50 & (65.1\%, 29\%) & (65.1\%, 28.3\%) & (\textbf{65.2\%}, 32.2\%) & (65\%, \textbf{32.7\%}) \\
 \cline{2-7}
\end{tabular}
\caption{\centering \small \feddynat performance on VGG model for communication budgets of 20 and 50 rounds. We evaluate \fedavgat and \fedcurvat with $E=\{1, 5, 20\}$ and report the best individual performance.}
\label{FedDynAT_VGG_100rounds}
\end{table}

\subsubsection{Convergence Speed for \feddynat}

We present the number of rounds needed for convergence with \feddynat compared to \fedavgat evaluated with $E=1$. The results for NIN model  are in Table~\ref{FedDynAT_NIN_convergence} , for the EMNIST-M model in Table~\ref{FedDynAT_EMNIST_convergence}, and for VGG in Table~\ref{FedDynAT_VGG_convergence}.

\begin{figure}[!ht]
\centering
\includegraphics[scale=0.5]{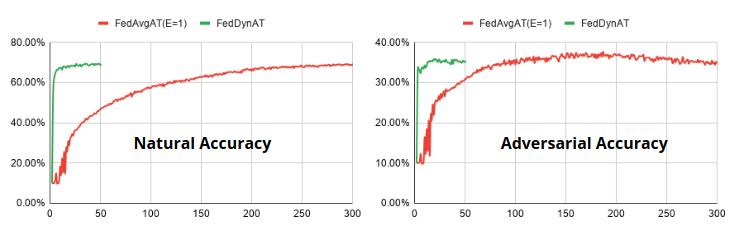} 
    \caption{\centering \small Accuracy vs communication round plot for \feddynat and \fedavgat evaluated with E=1, on $K=5$ client non-iid data split with VGG.}
    \label{fig:convergence_plot_fedavg_E=1_FedDynAT_VGG}
\end{figure}

The corresponding convergence plots for NIN and EMNIST-M networks are discussed in Section~\ref{convergence_speed_section}, and for VGG network in Figure~\ref{fig:convergence_plot_fedavg_E=1_FedDynAT_VGG}. 

Observe that FedDynAT converges
in significantly lower number of communication rounds as compared to \fedavgat(E=1), without any substantial drop in performance. 

\begin{table}[!ht]
\centering
\small
\setlength{\tabcolsep}{3pt} 
\begin{tabular}{p{1pt}cccc}
\multicolumn{4}{c}{} \\
\cline{2-5}
 &  & & \feddynat & \fedavgat(E=1) \\ 
 \cline{2-5}
&  K=2 &  Test Accuracy & (70.4\%, 38\%) & (\textbf{70.5}\%, \textbf{39.5}\%) \\ [-0.05in]
&&  Communication Rounds & \textbf{100} & 300 \\
 \cline{2-5}
&  K=5 &  Test Accuracy & (66.8\%, 35.9\%) & (\textbf{67.5}\%, \textbf{36}\%) \\ [-0.05in]
&&  Communication Rounds & \textbf{100} & 800 \\
 \cline{2-5}
 &  K=10 &  Test Accuracy & (\textbf{62.6}\%, \textbf{33.3}\%) & (52.7\%,31.2\%) \\ [-0.05in]
&&  Communication Rounds & \textbf{100} & 1400 \\
 \cline{2-5}
\end{tabular}
\caption{\centering \small The convergence accuracy and the number of communication rounds to convergence are listed here for the NIN model using \feddynat, and \fedavgat evaluated with $E=1$. \fedavgat(E=1) for 10 clients has not converged till 1400 rounds, hence the low accuracy.}
\vspace{-0.1in}
\label{FedDynAT_NIN_convergence}
\end{table}

\begin{table}[!ht]
\centering
\small
\setlength{\tabcolsep}{3pt} 
\begin{tabular}{p{1pt}cccc}
\multicolumn{4}{c}{} \\
\cline{2-5}
 &  & & \feddynat & \fedavgat(E=1) \\
 \cline{2-5}
& K=2 & Test Accuracy & (84.5\%, 64.5\%) & (84.5\%, 64.5\%) \\ [-0.05in]
&& Communication Rounds & \textbf{20} & 40 \\
 \cline{2-5}
& 5 & Test Accuracy & (83.5\%, 58\%) & (83.5\%, 58\%) \\ [-0.05in] 
&& Communication Rounds & \textbf{50} & 150 \\
\cline{2-5}
& 10 & Test Accuracy & (82.5\%, 56.5\%) & (82.5\%, 56.5\%) \\ [-0.05in] 
&& Communication Rounds & \textbf{100} & 250 \\
 \cline{2-5}
\end{tabular}
\caption{\centering \small The convergence accuracy and the number of communication rounds to convergence are listed here for EMNIST-M using \feddynat, and \fedavgat evaluated with $E=1$.}
\vspace{-0.1in}
\label{FedDynAT_EMNIST_convergence}
\end{table}

\begin{table}[!ht]
\centering
\small
\setlength{\tabcolsep}{3pt} 
\begin{tabular}{p{1pt}cccc}
\multicolumn{4}{c}{} \\
\cline{2-5}
 &  & & \feddynat & \fedavgat(E=1) \\ 
 \cline{2-5}
&  K=2 &  Test Accuracy & (70.4\%, \textbf{39}\%) & (\textbf{70.5}\%, \textbf{39}\%) \\ [-0.05in]
&&  Communication Rounds & \textbf{50} & 200 \\
 \cline{2-5}
&  K=5 &  Test Accuracy & (\textbf{69.1}\%, 35.6\%) & (69\%, \textbf{35.7}\%) \\ [-0.05in]
&&  Communication Rounds & \textbf{50} & 300 \\
 \cline{2-5}
 &  K=10 &  Test Accuracy & (\textbf{65}\%, 32.7\%) & (57.6\%,\textbf{33.9\%}) \\ [-0.05in]
&&  Communication Rounds & \textbf{50} & 900 \\
 \cline{2-5}
\end{tabular}
\caption{\centering \small The convergence accuracy and the number of communication rounds to convergence are listed here for the NIN model using \feddynat, and \fedavgat evaluated with $E=1$. \fedavgat(E=1) for 10 clients has not converged till 900 rounds, hence the low accuracy.}
\vspace{-0.1in}
\label{FedDynAT_VGG_convergence}
\end{table}

\end{document}